\documentclass[10pt,twocolumn,letterpaper]{article}

\usepackage{wacv}              

\usepackage{graphicx}%
\usepackage{multirow}%
\usepackage{amsmath,amssymb,amsfonts}%
\usepackage{amsthm}%
\usepackage{mathrsfs}%
\usepackage[title]{appendix}%
\usepackage{colortbl}
\usepackage{xcolor}%
\usepackage{textcomp}%
\usepackage{manyfoot}%
\usepackage{booktabs}%
\usepackage{array}
\usepackage{arydshln}
\usepackage{algorithm}%
\usepackage{algorithmicx}%
\usepackage{algpseudocode}%
\usepackage{listings}
\usepackage{wrapfig}
\usepackage{url}
\usepackage{multirow}
\usepackage{bbding}
\usepackage{pifont}
\usepackage{nicefrac}  
\usepackage{microtype}
\usepackage{url}
\usepackage{xcolor}
\newcommand{\cmark}{\ding{51}}
\newcommand{\xmark}{\ding{55}}

\usepackage{adjustbox}
\definecolor{ourslight}{RGB}{235,245,255}   
\definecolor{oursdark}{RGB}{210,230,250}    

\definecolor{best}{RGB}{220,245,220}        
\definecolor{second}{RGB}{255,235,210} 

\definecolor{wacvblue}{rgb}{0.21,0.49,0.74}
\usepackage[pagebackref,breaklinks,colorlinks,allcolors=wacvblue]{hyperref}

\def\wacvPaperID{2268} 
\def\confName{WACV}
\def\confYear{2026}

\title{DeepSTEP: Simultaneous Tracking and Estimation of Pose for Animals and Humans}

\author{
\begin{tabular}{ccc}
\begin{tabular}{c}
Shashikant Verma\\
Indian Institute of Technology\\
Gandhinagar, India
\end{tabular}
&
\begin{tabular}{c}
Harish Katti\\
National Institute of Health\\
Bethesda, USA
\end{tabular}
&
\begin{tabular}{c}
Soumyaratna Debnath\\
Indian Institute of Technology\\
Gandhinagar, India
\end{tabular}
\\[10pt]
\multicolumn{3}{c}{
\begin{tabular}{cc}
\begin{tabular}{c}
Yamuna N. Swamy\\
National Institute of Health\\
Bethesda, USA
\end{tabular}
&
\begin{tabular}{c}
Shanmuganathan Raman\\
Indian Institute of Technology\\
Gandhinagar, India
\end{tabular}
\end{tabular}
}
\vspace{0.4cm}
\end{tabular}
\\
shashikant.verma@iitgn.ac.in,
kattih2@nih.gov,
debnathsoumyaratna@iitgn.ac.in\\
yamuna.narayanaswamy@nih.gov,
shanmuga@iitgn.ac.in
}

\begin{document}
\maketitle
\begin{abstract}
Maintaining a consistent identity across video frames while estimating pose is crucial for downstream tasks such as behavior analysis and action recognition, yet existing pipelines largely decouple tracking and pose estimation. Top-down methods rely on per-frame detections followed by identity association, leading to increased computational cost and error propagation, whereas bottom-up approaches often struggle to preserve identity reliably in multi-instance settings.
To address these limitations, we introduce DeepSTEP (STEP), a Transformer-based framework for joint tracking and pose estimation across diverse animal species and humans. STEP unifies localization, identity maintenance, and keypoint estimation into a single model, eliminating the need for per-frame detections or post hoc matching. Our approach leverages discriminative model weight prediction to propagate target identity across frames while predicting anatomically meaningful keypoints.
Discriminative model prediction typically requires predefined target states. In contrast, STEP assumes only an initial bounding box in the first frame, removing the need for explicit keypoint initialization. This is enabled by two components: Gaussian Map Soft Prediction (GMSP) and Offset Map Regression Adapter (OMRA), which allow direct keypoint prediction and simplify inference. Starting from the initial bounding box, STEP tracks the subject and estimates its pose in subsequent frames.
We train and evaluate STEP on datasets spanning multiple species, demonstrating strong performance and greater efficiency than existing methods. By jointly modeling tracking and pose, STEP provides an effective and scalable solution.
\end{abstract}
    
\vspace{-0.5cm}
\section{Introduction}
\label{sec:intro}

Pose estimation is a fundamental and demanding task in computer vision, with diverse practical applications in the real world. When applied to video sequences, pose estimation proves invaluable across a range of applications, including action recognition, behavioral analysis, robotics, and surveillance, demonstrating its wide utility. Analyzing poses across videos involves tackling specific challenges, such as object detection, precise frame-to-frame identity tracking, and pose keypoint estimation. 

Despite significant progress, existing approaches typically address these components in a decoupled manner, solving object localization, pose estimation, and instance identity association in separate stages. Such designs introduce inefficiencies and are prone to error accumulation, particularly in video settings where consistent identity across frames must be maintained.
This decoupling is inherent to the dominant pose estimation paradigms, namely top-down and bottom-up approaches. Top-down methods \cite{xu2022vitpose, sun2019deep} first detect object instances and then estimate pose for each detection, requiring an additional step to associate identities across frames. In contrast, bottom-up approaches \cite{pishchulin2016deepcut, ye2023distilpose, insafutdinov2016deepercut} first predict all keypoints and subsequently group them into instances, which makes identity assignment inherently ambiguous in multi-instance scenarios. In both cases, pose estimation and identity reasoning are not jointly modeled, limiting their effectiveness for temporally consistent analysis.

To address these limitations, we propose DeepSTEP, a novel Transformer-based deep learning framework for \textbf{S}imultaneous \textbf{T}racking and \textbf{E}stimation of \textbf{P}ose, hereafter referred to as STEP.
Unlike existing pipelines that decouple pose estimation from identity association, STEP jointly models localization, identity propagation, and keypoint prediction within a unified architecture, leveraging discriminative model weight prediction to maintain target identity across frames while estimating pose.


In summary, our contributions are: 
\begin{itemize}
    \item We propose a unified framework for tracking and pose estimation that jointly models localization, identity propagation, and keypoint prediction, eliminating the need for per-frame detections and post-hoc identity association.
    
    \item We extend the discriminative model prediction paradigm to pose estimation by introducing Gaussian Map Soft Prediction (GMSP) and Offset Map Regression Adapter (OMRA), enabling keypoint prediction without explicit keypoint initialization.
    
    \item We design a temporal inference mechanism with confidence-based memory updates to maintain consistent identity and improve robustness across frames.
\end{itemize}

\section{Related Works}
\label{sec:rw}
Our proposed approach is inspired by established discriminative model prediction paradigms used in visual object tracking \cite{bhat2019learning} \cite{danelljan2019atom} \cite{dai2019visual} \cite{mayer2022transforming}. We explore the potential of this framework to predict keypoints for pose estimation while simultaneously performing tracking. In contrast to prevailing trends, which often focus on solving dedicated tasks sequentially, we demonstrate how vision transformer-based discriminative model weight predictions bridge the gap, enabling simultaneous predictions rather than sequential estimation. 

\noindent\textbf{Vision Transformers for Tracking.}
Modern trackers such as \cite{carion2020end} \cite{chen2021transformer} \cite{yu2021high}
utilize transformers to predict discriminative features for target localization and bounding-box regression.
These trackers involve a transformer encoder that encodes target-related features, followed by a transformer decoder that combines training and test features using cross-attention layers to compute discriminative features \cite{wang2021transformer} \cite{yu2021high} \cite{chen2021transformer}. 
Two specialized modules are employed by \cite{yu2021high} for localization and bounding box regression, while \cite{chen2021transformer} employs a feature fusion module. In \cite{zhang2021distractor}, a Multi-object tracker is introduced, drawing inspiration from tracking-by-detection. Additional works that share a similar approach, such as \cite{danelljan2019atom} and \cite{bhat2019learning}, incorporate a memory state to update online trackers.
Building on transformer-based formulations \cite{carion2020end, mayer2022transforming, mayer2024beyond}, recent works have explored model prediction frameworks that directly estimate task-specific weights for localization and regression. In a similar direction, \cite{verma2025gmot} investigates generic multiple-object tracking using a Mamba-based architecture.

\noindent\textbf{Vision Transformers for Pose Estimation.}
The evolution of pose estimation methods has swiftly shifted from CNNs \cite{xiao2018simple} to vision transformer-based networks, driven by their robust feature representations. 
In contrast to their use in discriminative model prediction-based tracking approaches, studies in pose estimation view transformers as more effective decoders \cite{li2021pose} \cite{li2021tokenpose} \cite{li2024efficient} \cite{debnath2025l3d} \cite{debnath2025computational}.
In \cite{yang2021transpose}, the transformer architecture processes features to capture global relationships within images. Introducing a token-based representation, \cite{li2021tokenpose} models relationships among pose keypoints. Additionally, HRFormer \cite{yuan2021hrformer} utilizes transformers to learn high-resolution features. These transformer-based pose estimation methods outperform widely used keypoint estimation benchmarks. 
Investigating the capabilities of straightforward vision transformers, \cite{xu2022vitpose} explores simplified transformer baselines that eliminate heavyweight CNN-based feature extractors and intricate transformer architectures found in existing neural models. More recently, pose estimation in challenging scenarios, such as low-light images and limited annotations, has been addressed in \cite{ai2024domain} and \cite{li2023scarcenet}.

\noindent\textbf{Pose Estimation and Tracking in videos.}
Object tracking and pose estimation in videos have extensive applications, including action recognition \cite{bouchrika2011using} and the analysis of human or animal behavior in various environments. Notably, markerless pose estimation is gaining popularity in neuroscience and the study of animal behaviors \cite{coria2022neurobiology}.
Expanding on the widely used CNN-based architecture of the DeepLabCut model \cite{mathis2018deeplabcut}, \cite{lauer2022multi} extends it to facilitate multi-animal tracking and pose estimation. SLEAP \cite{pereira2022sleap} employs a lightweight CNN, effectively tracking insects (Drosophila, Bees) and rodents. SuperAnimal \cite{ye2024superanimal} introduces a foundation model for cross-species pose estimation, enabling strong generalization across several animal species. In these approaches, pose estimation occurs frame-by-frame, whereas the tracking framework relies on a computationally intensive optical-flow-based method.
An encoder-decoder setup is proposed in \cite{sun2022self} to encode input frames and train the model to reconstruct spatiotemporal differences between video frames for keypoint discovery. Several studies have concentrated on analyzing human poses across videos. Approaches such as \cite{xiao2018simple}, \cite{yang2021transpose}, and \cite{yuan2021hrformer} involve framewise detection followed by dedicated modules for identification across frames. 

\section{Methodology}
\label{sec:meth}
This work introduces STEP, a Transformer-based target-model prediction network that simultaneously tracks and estimates pose by predicting keypoints.
We first provide a concise overview of existing Optimization and Transformer-based model predictors and then elaborate on our proposed approach, STEP, outlining its network architecture, training, and inference procedures in detail.

\begin{figure*}[t]
    \centering
    \includegraphics[width=\linewidth ]{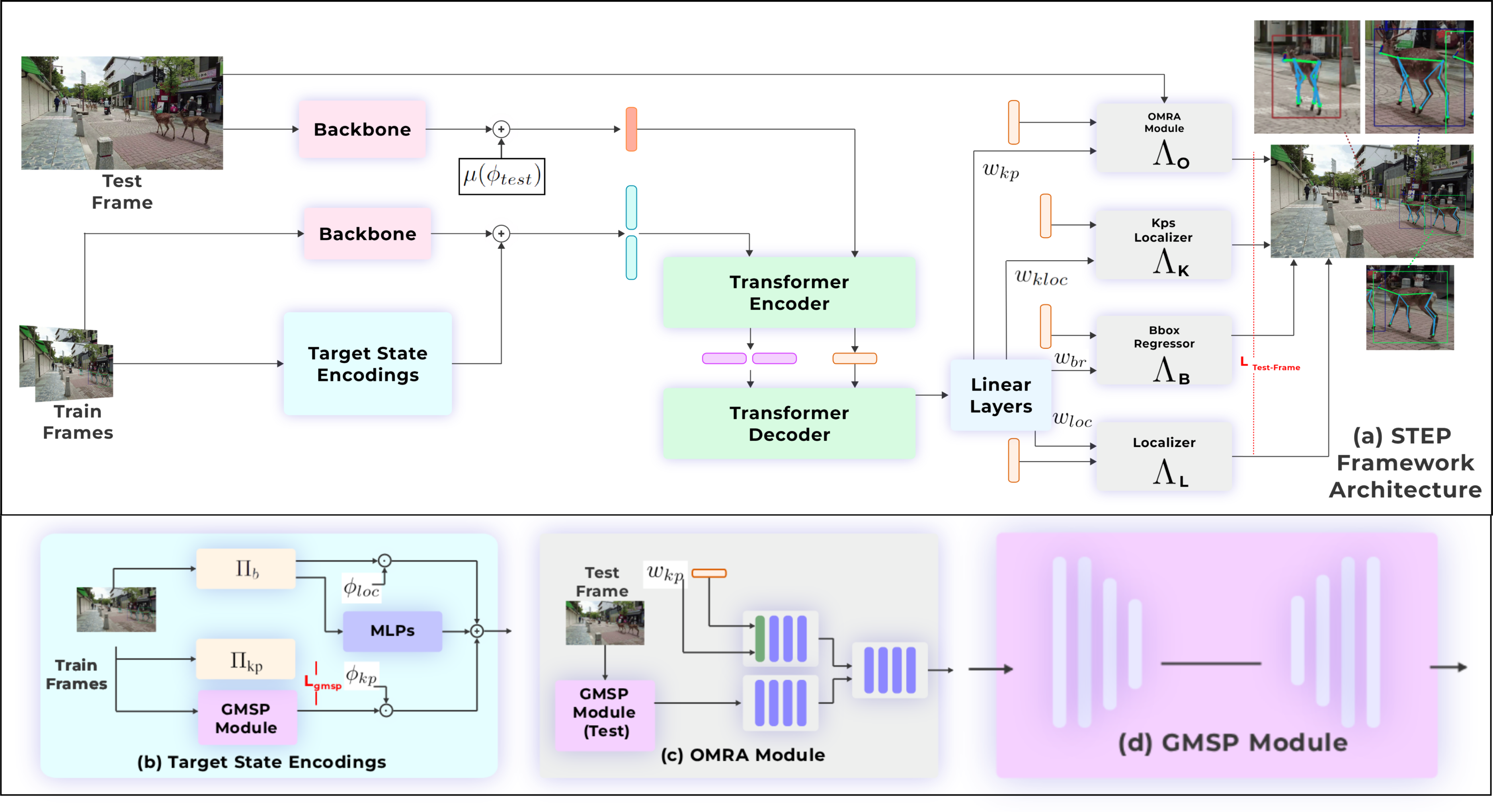}
    \vspace{-0.5cm}
    \caption{
    The proposed STEP framework for simultaneous tracking and pose estimation.
    (a) the full architecture, (b) the calculation of Target State Encodings, (c) the architecture of the offset map regression adapter (OMRA) module, and (d) Gaussian Soft Map Predictor (GMSP) Module.}
    \label{fig:arch}
\end{figure*}
\noindent \textbf{Background}.
Existing discriminative models like \cite{bhat2019learning} \cite{danelljan2019atom} \cite{dai2019visual} \cite{zheng2020learning} typically adhere to a shared foundational framework. They aim to solve optimization problems, as presented in Equation \ref{eq:taropt}, where the target model $\mathcal{T}(.)$ generates the intended target states.
\begin{equation}
    w = \arg\min_{\hat{w}} \sum_{(x,y) \in \mathcal{S}_{train}} \phi(\mathcal{T}(\hat{w};x),y) + \lambda r(\hat{w})
    \label{eq:taropt}
\end{equation}

\noindent Given a collection of training samples comprising a total of $m$ training frames, denoted as $\mathcal{S}_{train} = \{(x_i,y_i)\}_{i=1}^m$, where $x_i \in \mathcal{X}$ represents the high-dimensional feature map of the $i$-th frame and $y_i \in \mathcal{Y}$ represents the corresponding desired target state. 
The optimization problem outlined in Equation \ref{eq:taropt} aims to attain the optimal weight $w$ for the target model $\mathcal{T}$. This is achieved by minimizing residuals computed by the function $\phi(.)$, which measures the error between the output of the target model and the ground truth labels $y$. $r(.)$ is a regularization term weighted by a scalar $\lambda$.
Approaches following this optimization strategy optimize weights with limited information, relying solely on the provided training frames. Furthermore, they struggle to incorporate learned priors into the target model. 
Rather than directly minimizing the objective outlined in Equation \ref{eq:taropt}, \cite{mayer2022transforming} introduces a Transformer-based model predictor approach. This method predicts target model weights solely from data through end-to-end learning. 
Nevertheless, their primary emphasis lies in learning features that effectively differentiate the target state from previously encountered states within $\mathcal{S}_{train}$. For pose estimation, the model predictor should precisely capture information relevant to anatomical keypoints rather than focusing solely on localization tasks.

\subsection{Network Architecture}
\label{sec:na}
Our proposed network adopts an end-to-end learning approach to predict target-specific model weights directly from data. The overall pipeline is illustrated in Figure~\ref{fig:arch}.
The central idea behind our design is to move away from learning a single global predictor for all objects, and instead dynamically generate a target-conditioned model for each instance. This is particularly important for pose estimation, where the appearance and spatial configuration of keypoints can vary significantly across objects. By conditioning the model on the target instance, the network can better capture fine-grained, anatomically meaningful structures.
To this end, we build on the discriminative model-prediction framework introduced in~\cite{mayer2022transforming}. In contrast to conventional approaches that directly regress outputs, this paradigm learns to predict the parameters of task-specific heads conditioned on the target. This enables stronger target specificity and improves robustness to appearance variations.

We extend this formulation to the keypoint regression setting. Unlike~\cite{mayer2022transforming}, which primarily focuses on bounding box localization, our formulation explicitly incorporates information from $k$ keypoints associated with the target object. This extension is non-trivial for two reasons. First, keypoint estimation requires modeling structured spatial relationships beyond coarse object localization. Second, while bounding boxes can be easily initialized (e.g., via manual annotation or object detectors), obtaining keypoint annotations for initialization is significantly more cumbersome and often impractical.
To address this challenge, we design our framework to operate without explicit keypoint initialization at inference time. Instead, the network learns to automatically infer target-specific keypoint structure solely from bounding-box supervision. This allows the model to retain the simplicity of bounding-box-based initialization while still enabling accurate and instance-specific pose estimation.

\noindent \textbf{Overview.}
Given an input frame, we first extract high-dimensional feature maps using a pre-trained ImageNet backbone~\cite{deng2009imagenet}. These features serve as a generic representation of the scene. We then augment them with \emph{target-state encodings}, which inject instance-specific spatial information (e.g., bounding-box and keypoint cues).
The combined representations from training frames, along with features from the test frame, are processed by a Transformer encoder-decoder. The Transformer models interactions between target-specific cues and global image context, and predicts a set of target-conditioned model weights. These weights are subsequently used by dedicated regression modules to perform localization, bounding box estimation, and keypoint prediction.

The main components of our network are described in detail in the following subsections.

\subsubsection{Target State Encodings} 
A key component of our framework is the representation of target-specific information in a form that can guide both tracking and pose estimation. While standard tracking formulations focus on object localization and bounding box regression, pose estimation additionally requires modeling the spatial configuration of multiple keypoints.
We represent the target state using two complementary signals: \emph{Gaussian maps} and \emph{offset maps}. The Gaussian maps provide a coarse localization cue by highlighting regions likely to contain the target or its keypoints, whereas the offset maps encode fine-grained spatial displacements from each feature location to the precise target positions. This decomposition allows the model to first localize candidate regions and then refine them accurately.

Let $K^t \in \mathbb{R}^{2 \times k}$ denote the $k$ keypoints of the target object, where each keypoint is given by $K_i^t = (x_i, y_i)$. Given an input image $I \in \mathbb{R}^{H_i \times W_i \times 3}$, a backbone network extracts feature maps $X \in \mathbb{R}^{H_f \times W_f \times n}$ with downsampling factor $s$.

\noindent \textbf{Global vs. Target-specific keypoint maps.}
We distinguish between two types of Gaussian keypoint representations. First, we construct a global keypoint map $\mathrm{K}_{\text{gmsp}} \in \mathbb{R}^{H_f \times W_f \times k}$, where each channel aggregates Gaussian activations from all annotated objects in the frame for a given keypoint type. This representation captures category-level spatial priors and is used to train the GMSP module to predict potential keypoint locations in a class-agnostic manner.
In contrast, for the target object, we derive a target-specific keypoint map $\mathrm{K}^t_{\text{gm}}$ by retaining only the Gaussian responses corresponding to the object of interest and suppressing all others. This separation allows the GMSP module to first learn a soft distribution of plausible keypoint locations across all objects, which is subsequently refined into target-specific predictions by the downstream Transformer and regression modules. While Gaussian maps capture coarse localization cues, precise keypoint estimation requires modeling fine-grained spatial displacements. To this end, we compute offset maps specifically for the target object. For each spatial location $(j^x, j^y)$ on the feature map, we compute keypoint offset representations:
\begin{equation}
\kappa_i = \left( \left\lfloor\dfrac{s}{2}\right\rfloor + s \cdot j^x - x_i,\;
\left\lfloor\dfrac{s}{2}\right\rfloor + s \cdot j^y - y_i \right),
\end{equation}
resulting in a dense offset map $\mathrm{K}^t_{\text{om}} \in \mathbb{R}^{H_f \times W_f \times 2k}$.

\noindent \textbf{Bounding Box Maps.} Similarly, for a bounding box $\beta = \{b_{x_1}, b_{y_1}, b_{x_2}, b_{y_2}\}$, we adopt the widely used LTRB representation~\cite{mayer2022transforming,yu2021high} to construct dense offset maps. For each spatial location $(j^x, j^y)$ on the feature map, the bounding box is translated into an offset representation as:
$\mathrm{B}_{\text{om}} = \{ \lfloor(s/2)\rfloor + s \cdot j^x- b \} \;\; \forall \;b \in \beta$
resulting in $\mathrm{B}_{\text{om}} \in \mathbb{R}^{H_f \times W_f \times 4}$. 
In addition, we generate a Gaussian map $\mathrm{B}_{\text{gm}}$ centered at the bounding box center, which serves as a localization cue for the target. 

\noindent \textbf{Encoding.} 
To encode this target information into the feature space, we use learnable projections and embeddings. Specifically, we introduce embeddings $\phi_{\text{loc}} \in \mathbb{R}^{1 \times n}$ and $\phi_{\text{kp}} \in \mathbb{R}^{k \times n}$ to modulate localization and keypoint cues. The resulting training features are defined as:
\begin{equation}
f_i = X + \Psi_b(B_{\text{om}}) + \Psi_{\text{kp}}(K^t_{\text{om}})
+ \phi_{\text{loc}} \cdot B_{\text{gm}} + \phi_{\text{kp}} \cdot K^t_{\text{gm}},
\end{equation}
where $\Psi$ denotes learnable mappings and $\cdot$ represents element-wise modulation.

\noindent \textbf{Keypoint-Free Encoding.}
The formulation above assumes access to target-specific keypoints for initialization in the reference frame. However, unlike bounding boxes, initializing keypoints is non-trivial in practice, as it requires either ground-truth annotations or manual user input, both of which are cumbersome and often unavailable during inference. 
To address this, we replace explicit keypoint supervision with a learned soft approximation. Specifically, we introduce a soft keypoint prior $\hat{K}{\text{gmsp}}$, which approximates Gaussian keypoint maps directly from appearance cues. This leads to the modified encoding:
\begin{equation}
f_i = X + \Psi_b(B_{\text{om}})
+ \phi_{\text{loc}} \cdot B_{\text{gm}}
+ \phi_{\text{kp}} \cdot \hat{K}_{\text{gmsp}}.
\end{equation}
This design enables the network to automatically infer keypoint structure while relying only on bounding-box initialization.
Further, we incorporate the test frame encoding, composed of deep features $X_{\text{test}}$ and a learnable token $\phi_{\text{test}}$, resulting in $f_{\text{test}} = X_{\text{test}} + \mu(\phi_{\text{test}})$. Here, $\phi_{\text{test}}$ represents the learnable token, and $\mu(\cdot)$ signifies the repetition of the token for each dimension of $X_{\text{test}}$.

\subsubsection{Transformer Encoder and Decoder}
The Transformer module serves as the core mechanism for generating target-conditioned model parameters. Given training features $\{f_i\}$ and test features $f_{\text{test}}$, the goal is to learn a mapping from target-specific context to a set of weights that can be directly applied for prediction on the test frame.
The encoder processes the training features $f_i$ to capture target-specific appearance and spatial structure, while the decoder conditions on the test features $f_{\text{test}}$ to adapt these representations to the current frame. This interaction enables the model to transfer target information across frames while accounting for appearance changes and motion.
The Transformer predicts a set of target-specific model weights, including localization weights ($w_{\text{loc}}$, $w_{\text{kloc}}$) and regression weights ($w_{\text{br}}$, $w_{\text{kp}}$) for bounding boxes and keypoints, respectively. These weights parameterize lightweight task-specific heads used for dense prediction.
Denoting the encoded representations as $z_i$ and $z_{\text{test}}$, the predicted weights act as dynamic filters applied on $z_{\text{test}}$, allowing the model to perform instance-specific inference. This formulation effectively decouples representation learning from prediction, enabling stronger target specificity compared to fixed-weight architectures.

\subsubsection{Regression Modules}
\label{sec:regm}

The regression modules translate the target-conditioned representations into final predictions for localization, bounding box estimation, and keypoint regression. To enable keypoint estimation without explicit initialization, we combine global keypoint priors with target-specific adaptation through two components: the Gaussian Map Soft Prediction (GMSP) module and the Offset Map Regression Adapter (OMRA).

\noindent \textbf{GMSP Module.}
The GMSP module predicts a soft keypoint map $\hat{K}_{\text{gmsp}}$ that captures plausible locations of all keypoints in a class-agnostic manner. It is trained only on training frames and remains fixed during inference, serving as a global structural prior over keypoint configurations.

\noindent \textbf{Offset Map Regression Adapter (OMRA).}
The OMRA module, denoted as $\Lambda_O$, refines the global keypoint prior into target-specific predictions. It combines (i) target-conditioned responses obtained by applying the predicted weights $w_{\text{kp}}$ to encoded test features $z_{\text{test}}$, and (ii) the soft maps from the GMSP module. The fused representation is used to predict the target-specific keypoint offset map $\hat{K}^t_{\text{om}}$.

\noindent \textbf{Bounding Box and Localization Heads.}
We define the bounding box regressor $\Lambda_B$, box localizer $\Lambda_L$, and keypoint localizer $\Lambda_K$ as lightweight prediction heads operating on encoded test features. Each head applies the corresponding predicted weights ($w_{\text{loc}}$, $w_{\text{kloc}}$, $w_{\text{br}}$), followed by convolutional refinement. These modules predict the bounding box Gaussian map $\hat{B}_{\text{gm}}$, keypoint Gaussian maps $\hat{K}^t_{\text{gm}}$, and bounding box offset maps $\hat{B}_{\text{om}}$.

\begin{table*}[t]
\scriptsize
\setlength{\tabcolsep}{3pt}
\caption{Comparison with state-of-the-art pose estimation methods, including TransPose \cite{yang2021transpose}, YOLOv8 \cite{jocher2023yolo}, ViPNAS \cite{xu2021vipnas}, ViTPose \cite{xu2022vitpose}, DeepLabCut \cite{lauer2022multi}, SuperAnimal \cite{ye2024superanimal}, RTMPose \cite{rtmpose}, HRNet \cite{sun2019deep}, and PAVENet \cite{yu2026end}. The best and second-best results are highlighted in green and orange, respectively. For the tracking performance of STEP, please refer to Figure \ref{fig:tracking}.
}
\resizebox{\textwidth}{!}{
\begin{tabular}{lcccccccccccc}
\toprule

\multirow{2}{*}{\textbf{Model}} &
\multicolumn{4}{c}{\textbf{APT36K}} &
\multicolumn{4}{c}{\textbf{APT10K}} &
\multicolumn{4}{c}{\textbf{CrowdPose}} \\

\cmidrule(lr){2-5}
\cmidrule(lr){6-9}
\cmidrule(lr){10-13}

& MSE $\downarrow$ & OKS $\uparrow$ & PDJ@0.08 & PDJ@0.05
& MSE $\downarrow$ & OKS $\uparrow$ & PDJ@0.08 & PDJ@0.05
& MSE $\downarrow$ & OKS $\uparrow$ & PDJ@0.08 & PDJ@0.05 \\

\midrule

TransPose & 530.61 & 0.749 & 0.747 & 0.641 & 732.52 & 0.548 & 0.621 & 0.492 & -- & -- & -- & -- \\

YOLOv8-M  & 459.49 & 0.798 & 0.846 & 0.715 & 423.89 & 0.567 & 0.643 & 0.544 & -- & -- & -- & -- \\

YOLOv8-L  & 449.52 & 0.756 & 0.856 & 0.725 & 406.56 & 0.567 & 0.656 & 0.516 & -- & -- & -- & -- \\

ViPNAS    & 596.66 & 0.757 & 0.729 & 0.638 & 647.12 & 0.503 & 0.671 & 0.503 & -- & -- & -- & -- \\

ViTPose   & 490.20 & 0.804 & 0.916 & \cellcolor{second}0.814
          & 548.07 & 0.562 & 0.692 & 0.572
          & 339.76 & 0.681 & 0.782 & 0.645 \\

SuperAnimal   & \cellcolor{best}382.08 & \cellcolor{best}0.835 & \cellcolor{second}0.938 & \cellcolor{best}0.846
          & \cellcolor{best}360.34 & \cellcolor{best}0.598 & \cellcolor{second}0.690 & \cellcolor{best}0.583
          & -- & -- & -- & -- \\

RTMPose   & -- & -- & -- & -- & -- & -- & -- & -- & \cellcolor{second}241.07 & 0.715 & \cellcolor{second}0.887 & \cellcolor{second}0.723 \\

HRNet     & -- & -- & -- & -- & -- & -- & -- & -- & 371.41 & 0.653 & 0.763 & 0.625 \\

PaveNet     & -- & -- & -- & -- & -- & -- & -- & -- & 296.37 & 0.691 & 0.815 & 0.642 \\

Ours      & 442.76 & 0.830 & 0.936 & 0.799
          & 482.69 & 0.571 & 0.682 & 0.565
          & 292.86 & \cellcolor{second}0.734 & 0.861 & 0.702 \\

Ours*     & \cellcolor{second}404.37 & \cellcolor{second}0.832 & \cellcolor{best}0.946 & 0.804
          & \cellcolor{second}385.57 & \cellcolor{second}0.586 & \cellcolor{best}0.694 & \cellcolor{second}0.582
          & \cellcolor{best}214.87 & \cellcolor{best}0.736 & \cellcolor{best}0.891 & \cellcolor{best}0.728 \\

\midrule

\multicolumn{4}{c}{\textbf{\quad \quad \quad \quad \quad \quad \quad \quad \quad \quad \quad \quad TriMouse}} &
\multicolumn{4}{c}{\textbf{\quad \quad \quad \quad \quad \quad \quad \quad \quad \quad \quad \quad  Fish}} &
\multicolumn{4}{c}{\textbf{\quad \quad \quad \quad \quad \quad \quad \quad \quad \quad \quad \quad Marmoset}} \\

\cmidrule(lr){2-5}
\cmidrule(lr){6-9}
\cmidrule(lr){10-13}

TransPose   & 119.15 & 0.885 & 0.925 & 0.886 & 99.33 & 0.954 & 0.913 & 0.801 & 223.64 & 0.734 & 0.796 & 0.714 \\

YOLOv8-M    & 82.34 & 0.901 & 0.912 & 0.860 & 71.15 & 0.952 & 0.925 & 0.823 & 166.31 & 0.763 & 0.830 & 0.744 \\

YOLOv8-L    & \cellcolor{second}65.18 & \cellcolor{second}0.903 & 0.928 &\cellcolor{best} 0.918
            & 59.58 & 0.952 & \cellcolor{second}0.943 & \cellcolor{second}0.836
            & 164.56 & 0.763 & 0.839 & \cellcolor{second}0.792 \\

DeepLabCut  & 102.76 & 0.897 & 0.923 & 0.885 & 92.12 & 0.958 & 0.929 & 0.819 & 190.47 & 0.768 & 0.829 & 0.727 \\

SuperAnimal  & 73.21 & 0.902 & 0.943 & 0.905 & -- & -- & -- & -- & -- & -- & -- & -- \\

Ours        & 75.96 & 0.902 & \cellcolor{second}0.946 & 0.916
            & \cellcolor{second}54.24 & \cellcolor{second}0.959 & 0.931 & \cellcolor{best}0.842
            & \cellcolor{second}161.95 & \cellcolor{second}0.779 & \cellcolor{second}0.859 & 0.784 \\

Ours*       & \cellcolor{best}67.35 & \cellcolor{best}0.904 & \cellcolor{best}0.948 & \cellcolor{second}0.915
            & \cellcolor{best}47.59 & \cellcolor{best}0.960 & \cellcolor{best}0.962 & 0.838
            & \cellcolor{best}159.68 & \cellcolor{best}0.784 & \cellcolor{best}0.858 & \cellcolor{best}0.799 \\

\bottomrule
\end{tabular}
}
\label{tab:sota}
\end{table*}

\subsection{Training and Inference}
\label{sec:tif}

\noindent \textbf{Training.}
We train the model using short video clips. Given a sequence 
$\mathcal{S} = \{s_0, \ldots, s_n\}$ with bounding box and keypoint annotations, 
we randomly sample two frames as training frames and one frame as the test frame.
For each training frame, target state encodings are computed as described in Section~\ref{sec:na}. These encodings are used by the Transformer to predict target-conditioned model weights, which are subsequently applied to the test frame for localization, bounding box regression, and keypoint estimation.
The prediction heads operate on the test frame as follows:
$\Lambda_O$ predicts $\hat{K}^t_{\text{om}}$, 
$\Lambda_K$ predicts $\hat{K}^t_{\text{gm}}$, 
$\Lambda_B$ predicts $\hat{B}_{\text{om}}$, and 
$\Lambda_L$ predicts $\hat{B}_{\text{gm}}$. 
These predictions are supervised using ground-truth maps 
$\{K^t_{\text{om}}, K^t_{\text{gm}}, B_{\text{om}}, B_{\text{gm}}\}$ constructed as described in Section~\ref{sec:na}.
To enforce a clear separation between global keypoint prior learning and target-specific prediction, the regression losses are computed only on the test frame, while the GMSP module is trained exclusively on the training frames to predict $\hat{K}_{\text{gmsp}}$. 

\noindent \textbf{Loss Functions.}
We employ a combination of classification and regression losses. Gaussian map predictions are supervised using a target classification loss $\mathcal{L}_{\text{cls}}$~\cite{bhat2019learning}. Bounding box regression is trained using generalized IoU loss $\mathcal{L}_{\text{giou}}$~\cite{rezatofighi2019generalized}, while keypoint offset regression uses a hinge mean-squared loss $\mathcal{L}_{\text{hom}}$, applied only to visible keypoints.

The training objective combines a test-frame loss and a GMSP loss, as defined in Equation \ref{eq:loss}. We use $\lambda_1 = \lambda_3 = \lambda_5 = 100$, 
$\lambda_2 = \lambda_4 = 10$, stride $s=16$, and map resolution $18 \times 18$.
\begin{align}
\mathcal{L}_{\text{test}}
&= \lambda_1 \mathcal{L}_{\text{cls}}(B_{\text{gm}}, \hat{B}_{\text{gm}})
+ \lambda_2 \mathcal{L}_{\text{giou}}(B_{\text{om}}, \hat{B}_{\text{om}}) \nonumber\\
&\quad + \lambda_3 \mathcal{L}_{\text{cls}}(K^t_{\text{gm}}, \hat{K}^t_{\text{gm}})
+ \lambda_4 \mathcal{L}_{\text{hom}}(K^t_{\text{om}}, \hat{K}^t_{\text{om}}).
\end{align}
\begin{equation}
\mathcal{L}_{\text{tot}} =
\mathcal{L}_{\text{test}} +
\lambda_5 \mathcal{L}_{\text{cls}}(K_{\text{gm}}, \hat{K}_{\text{gmsp}}).
\label{eq:loss}
\end{equation}

\noindent \textbf{Inference.}
At inference, the model is initialized using a bounding box in the first frame $s_0$, obtained either manually or via an external detector. We initialize the memory as $\mathcal{S}_{\text{train}} = \{s_0, s_0\}$.
For each subsequent frame, the Transformer predicts target-conditioned weights, which are applied by the heads $\Lambda$ to produce localization, bounding box, and keypoint predictions.
We maintain a dynamic memory of training frames, updated based on prediction confidence. A new frame is added when a sufficient number of keypoints and the target localization exceed a threshold $\tau_m$; otherwise, the memory remains unchanged, avoiding unnecessary recomputation.
Final predictions are obtained by first identifying peak locations in the Gaussian maps $\hat{K}^t_{\text{gm}}$ and $\hat{B}_{\text{gm}}$, and then decoding the corresponding offsets from $\hat{K}^t_{\text{om}}$ and $\hat{B}_{\text{om}}$ into image space using the stride $s$.

\noindent \textbf{Implementation details.}
We use the Adam optimizer with an initial learning rate of $10^{-4}$, decayed by a factor of $0.1$ after 50 epochs. During inference, memory is updated when at least $50\%$ of keypoints and the target localization exceed a confidence threshold $\tau_m = 0.6$. All experiments are conducted on NVIDIA RTX 4090 GPUs.

\section{Results and Discussion}
\label{sec:rnd}
We evaluate STEP across diverse benchmarks spanning multiple species, environments, and data modalities. These include APT36K~\cite{yang2022apt} (30 animal species, 36K video frames), APT10K~\cite{yu2021ap} (60 species, 10K images), the DeepLabCut benchmark datasets~\cite{deeplabcutDeepLabCutBenchmark,lauer2022multi} (Tri-mouse, Marmoset, and Fish), and CrowdPose~\cite{li2019crowdpose} for human pose estimation. Together, these datasets cover both in-the-wild and controlled laboratory settings, enabling a comprehensive evaluation of cross-species generalization and robustness.

\noindent \textbf{Experimental Setup.}
For image-based datasets, we synthesize short sequences using random affine transformations to enable temporal training. This allows the model to learn motion-consistent representations even in the absence of native video data. For evaluation, we generate smooth synthetic sequences of length 15 for validation splits, ensuring consistency across methods. The validation sets are organized as $(n_v, n_{tid}, n_f)$, where $n_v$ denotes the number of sequences, $n _ {tid}$ the number of unique tracking instances, and $ n_f$ the number of frames per sequence. Across APT36K, APT10K, CrowdPose, Fish, TriMouse, and Marmoset datasets, these configurations are $(237,348,15)$, $(201,269,15)$, $(1572,6803,15)$, $(24,336,15)$, $(24,78,15)$, and $(199,239,15)$.
We train separate models per dataset due to differences in keypoint definitions. This requires only minor modifications to the final layers of the regression heads, demonstrating STEP's flexibility in adapting to varying annotation schemes. These minor modifications enable efficient fine-tuning and facilitate the model's rapid adaptation to new datasets.

\subsection{Comparisons}
Table~\ref{tab:sota} compares STEP with state-of-the-art methods on pose estimation using Mean Squared Error (MSE), Object Keypoint Similarity (OKS), and Percentage of Detected Joints (PDJ). Here, PDJ@$x$ considers a keypoint correctly detected if it lies within a distance of $x \cdot d$, where $d$ is the diagonal of the target bounding box. We report two variants of our method: \textbf{Ours} (STEP), which performs joint tracking and pose estimation using only the initial frame bounding-box initialization, and \textbf{Ours*} (STEP*), which uses ground-truth bounding boxes for each frame following the standard top-down evaluation protocol. Per-keypoint error analysis for all models is included in the \textit{supplementary material}.
As shown in Table~\ref{tab:sota}, STEP is competitive with state-of-the-art methods on pose estimation benchmarks, while operating under a more challenging setting where no external detector is used. In the top-down setting, STEP* further improves performance and surpasses several state-of-the-art pose estimation methods across benchmarks, demonstrating the effectiveness of the proposed pose estimation module when provided with accurate localization. A key advantage of STEP is its ability to perform \emph{simultaneous tracking and pose estimation in a unified framework}, eliminating the need for a separate detection stage. In contrast to standard top-down pipelines that rely on per-frame detectors, this design enables significantly faster inference, achieving approximately 1.63 seconds per 100 frames compared to around 5 seconds per 100 frames for detector-based pipelines (Table~\ref{tab:fps}).

\begin{figure}[t]
    \centering
    \includegraphics[width=\columnwidth]{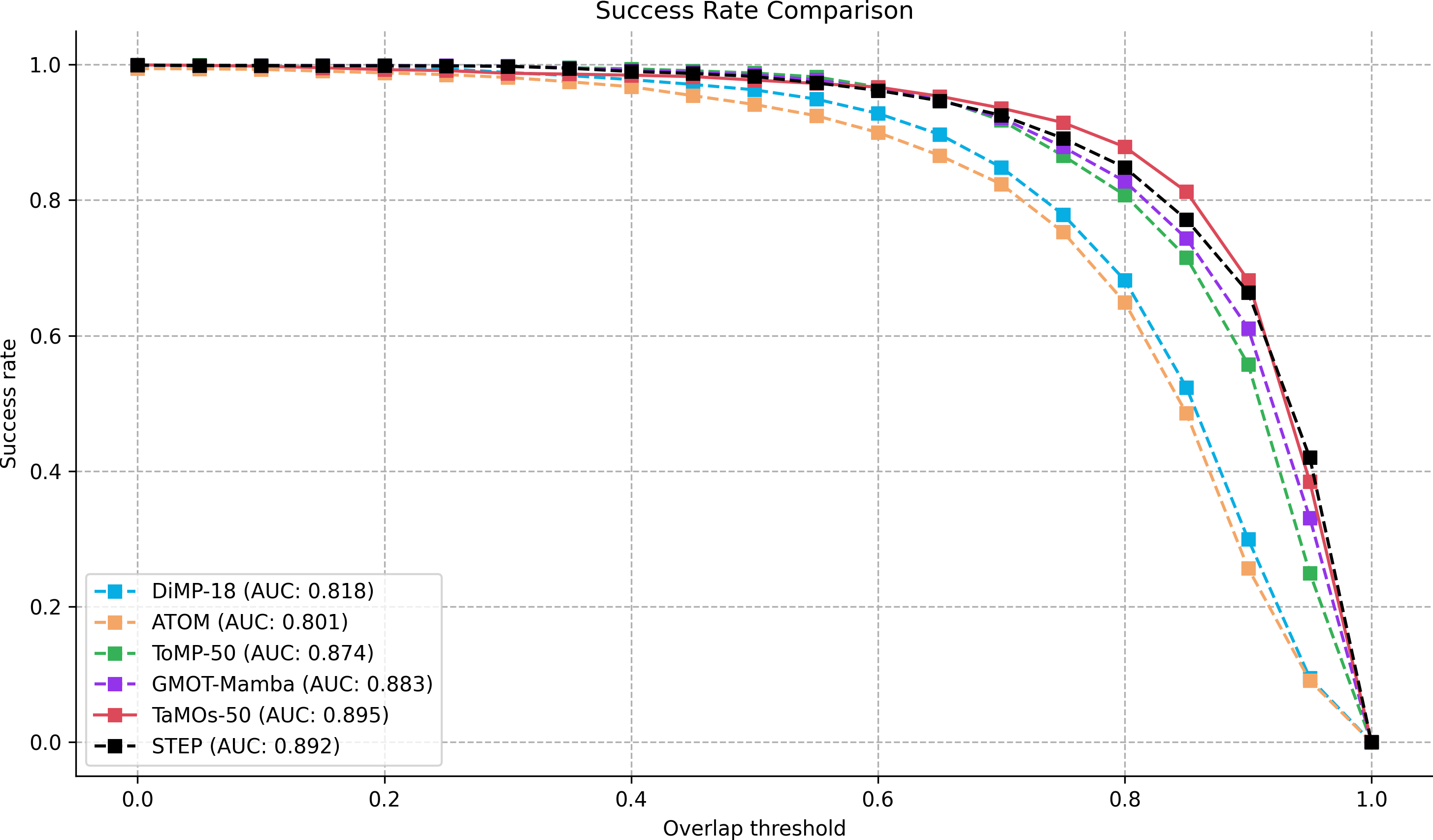}
    \vspace{-4pt}
    \caption{AUC and Overlap Precision $(\text{OP}_T)$ for bounding box tracking in STEP compared with: DiMP \protect\cite{bhat2019learning}, ToMP \protect\cite{mayer2022transforming}, ATOM \protect\cite{danelljan2019atom}, TAMOS \protect\cite{mayer2024beyond}, and GMOT-Mamba \protect\cite{verma2025gmot}.}
    \label{fig:tracking}
\end{figure}

This unified formulation is further validated in Figure~\ref{fig:tracking}, which evaluates tracking performance on APT36K using Overlap Precision (OP) across varying thresholds, reports the corresponding Area Under the Curve (AUC), and provides a visual comparison of tracking performance. STEP demonstrates tracking accuracy comparable to that of strong specialized trackers while maintaining pose estimation capability. Notably, STEP achieves an AUC of 0.894, close to state-of-the-art trackers, despite not being explicitly designed as a dedicated tracking model.
Overall, STEP offers a unified approach to tracking and pose estimation, achieving performance comparable to specialized pose pipelines while improving inference efficiency through a single joint formulation that removes the need for per-frame detection.

\begin{table}[t]
\scriptsize
\setlength{\tabcolsep}{8pt}
\renewcommand{\arraystretch}{1.1}
\begin{center}
\begin{tabular}{l c c c c}
\toprule
\textbf{Model} & \textbf{Task} & \textbf{FPS} & \textbf{Detector} & $T_{100}$ (s) \\
\midrule
\rowcolor{ourslight}
STEP  & Track + Pose & 63   & $\times$ & 1.63 \\
\rowcolor{oursdark}
STEP* & Pose            & $\sim$63 & $\checkmark$ & 6.13 \\

ViTPose \cite{xu2022vitpose}       & Pose & 231 & $\checkmark$ & 4.97 \\
TransPose \cite{yang2021transpose} & Pose & 43  & $\checkmark$ & 6.87 \\
YOLOv8-L \cite{jocher2023yolo}     & Pose & 113 & $\checkmark$ & 5.43 \\
YOLOv8-M \cite{jocher2023yolo}     & Pose & 161 & $\checkmark$ & 5.16 \\
RTMPose \cite{rtmpose}            & Pose & 263 & $\checkmark$ & 4.92 \\
DeepLabCut \cite{lauer2022multi}   & Pose & 68  & $\checkmark$ & 6.01 \\
HRNet \cite{sun2019deep}           & Pose & 31  & $\checkmark$ & 7.77 \\
ViPNAS \cite{xu2021vipnas}         & Pose & 117 & $\checkmark$ & 5.40 \\

\bottomrule
\end{tabular}
\end{center}
\vspace{-0.4cm}
\caption{Inference speed comparison.}
\label{tab:fps}
\end{table}

\noindent \textbf{Inference Speed Analysis.}
We evaluate inference speed on a standardized number of frames equivalent to the APT36K validation set (Table~\ref{tab:fps}). To ensure fair comparison, we follow established evaluation protocols for resolution and cropping across existing methods.
STEP achieves an average of 63 FPS while jointly tracking and estimating pose with only first-frame initialization. In contrast, STEP* and other top-down approaches incur additional computational overhead from per-frame detection. For reference, commonly used detectors such as MegaDetector~\cite{hernandez2024pytorchwildlife} and DINO~\cite{zhang2022dino} operate at 29 FPS and 22 FPS, respectively, creating a bottleneck in end-to-end pipelines.
We also report $T_{100}$, the time required to process 100 frames, including detection, to provide a holistic comparison. Notably, STEP inherently maintains target identity across frames, whereas other approaches require additional post-processing, such as tracklet stitching~\cite{lauer2022multi}. 
All experiments are conducted with a batch size of 1 on an NVIDIA RTX 4090 GPU.

\noindent \textbf{Model Complexity and Detector Dependency.} We include ViPNAS \cite{xu2021vipnas} (13.5M), TransPose \cite{yang2021transpose} (17.5M), DeepLabCut \cite{lauer2022multi} (25.6M), YOLOv8-M/L-Pose \cite{jocher2023yolo} (26.4M/44.4M), HRNet \cite{sun2019deep}(28.5M), and ViTPose \cite{xu2022vitpose} (86M) in our comparison. Most of these methods rely on accurate per-frame detections or bounding boxes, often requiring an additional detector, such as DINO \cite{zhang2022dino} (47M) or MegaDetector \cite{hernandez2024pytorchwildlife} (139.9M). In contrast, STEP (75M) requires a detector only for first-frame initialization and does not rely on repeated per-frame detections.

\subsection{Ablation Study}
\label{sec:abl}
We perform ablation studies on APT36K and APT10K. All variants are trained for 100 epochs and evaluated on a reduced validation split of 150 sequences per dataset.

\noindent \textbf{1. Natural vs Synthetic Video Data.}
Table~\ref{tab:all_ablation} compares models trained on natural and synthetically generated sequences, with each model evaluated on both the natural and synthetic validation splits. The model trained on natural sequences achieves the best overall performance across both validation settings. Nevertheless, when only image-based training data are available, STEP trained on synthetically generated sequences still generalizes effectively to real videos and achieves competitive performance. This suggests that STEP learns target-conditioned spatial relationships rather than relying solely on motion patterns in data.

\begin{table}[t]
\scriptsize
\setlength{\tabcolsep}{4pt}
\renewcommand{\arraystretch}{1.1}
\centering

\textbf{(a) Natural vs Synthetic Training}\par
\vspace{3pt}

\makebox[\linewidth]{%
\begin{tabular}{c ccc ccc}
\toprule

\multirow{2}{*}{\begin{tabular}{c}
\textbf{Model} \\
\textbf{Trained on}
\end{tabular}} &
\multicolumn{3}{c}{\textbf{Natural (APT36K)}} &
\multicolumn{3}{c}{\textbf{Synthetic (APT10K)}} \\

\cmidrule(lr){2-4}
\cmidrule(lr){5-7}

& MSE $\downarrow$ & OKS $\uparrow$ & PDJ $\uparrow$
& MSE $\downarrow$ & OKS $\uparrow$ & PDJ $\uparrow$ \\

\midrule

Synthetic & 374.28 & 0.635 & 0.749 & 395.76 & 0.764 & 0.818 \\ 

\rowcolor{ourslight}
Natural   & 356.68 & 0.733 & 0.872 & 397.98 & 0.791 & 0.861 \\ 

\bottomrule
\end{tabular}%
}

\vspace{6pt}

\textbf{(b) GMSP Ablation}\par
\vspace{3pt}

\makebox[\linewidth]{%
\begin{tabular}{c ccccc}
\toprule

\textbf{Model Config} & $\phi_{\text{kp}}$ & Init & MSE $\downarrow$ & OKS $\uparrow$ & PDJ $\uparrow$ \\

\midrule

\rowcolor{ourslight}
w GMSP   & \cmark & \xmark & 356.68 & 0.733 & 0.872 \\ 
w/o GMSP          & \cmark & \cmark & 464.78 & 0.698 & 0.832 \\ 
w GMSP            & \xmark & \xmark & 372.94 & 0.714 & 0.841 \\ 
w/o GMSP          & \xmark & \cmark & 468.78 & 0.628 & 0.802 \\ 

\bottomrule
\end{tabular}%
}

\vspace{6pt}

\textbf{(c) OMRA Ablation}\par
\vspace{3pt}

\makebox[\linewidth]{%
\begin{tabular}{c ccc}
\toprule

\textbf{Model Config} & MSE $\downarrow$ & OKS $\uparrow$ & PDJ $\uparrow$ \\

\midrule

\rowcolor{ourslight}
w OMRA              & 356.68 & 0.733 & 0.872 \\ 
w/o OMRA            & 378.52 & 0.702 & 0.821 \\ 

\bottomrule
\end{tabular}%
}

\vspace{6pt}

\textbf{(d) Training Frame Ablation}\par
\vspace{3pt}

\makebox[\linewidth]{%
\begin{tabular}{c cc cc}
\toprule

\textbf{\#Frames} & \multicolumn{2}{c}{\textbf{APT36K}} & \multicolumn{2}{c}{\textbf{APT10K}} \\

\cmidrule(lr){2-3}
\cmidrule(lr){4-5}

& MSE $\downarrow$ & OKS $\uparrow$
& MSE $\downarrow$ & OKS $\uparrow$ \\

\midrule

1 initial       & 465.48 & 0.697 & 536.69 & 0.684 \\ 
1 + 1 recent             & 360.87 & 0.723 & 397.30 & 0.768 \\ 
1 + 2 recent             & 358.31 & 0.723 & 395.82 & 0.778 \\ 
Rolling-2                & 495.89 & 0.682 & 543.49 & 0.673 \\ 
Rolling-3                & 492.62 & 0.674 & 538.95 & 0.686 \\ 

\rowcolor{ourslight}
Conf.-2                  & 356.68 & 0.733 & 395.76 & 0.764 \\ 
Conf.-3                  & 357.38 & 0.742 & 394.89 & 0.764 \\ 

\bottomrule
\end{tabular}%
}
\caption{Ablation studies across natural vs synthetic training data (a), STEP modules (b), and temporal memory update strategies (c).}
\label{tab:all_ablation}
\end{table}

\noindent \textbf{2. Architecture Ablations.}
Table~\ref{tab:all_ablation} (b) evaluates the contribution of the GMSP module, while (c) assesses the effectiveness of the OMRA module. Replacing GMSP with GT keypoints and removing its feature-injection branch results in a performance drop, validating that learned soft keypoint priors are more effective than explicit keypoint initialization. In addition to removing the need for keypoint initialization, GMSP provides richer contextual cues for keypoint localization.
We further evaluate the impact of target-state embeddings in Table~\ref{tab:all_ablation} (b). Incorporating $\phi_{\text{kp}}$ significantly improves pose estimation accuracy, demonstrating the importance of keypoint-aware modulation. Consistent with prior work~\cite{mayer2022transforming}, $\phi_{\text{loc}}$ also improves localization performance.

\noindent \textbf{3. Memory Update}
Table~\ref{tab:all_ablation} (d) evaluates the effect of updating the training set $\mathcal{S}_{\text{train}}$ during inference using confidence thresholding ($\tau_m$). Memory updates lead to consistent performance improvements by adapting the target representation over time, thereby improving robustness to appearance changes and mitigating drift. Qualitative illustrations in the supplementary material further demonstrate stable tracking and pose estimation under challenging conditions.

\subsection{Supplementary Material}
Additional results are provided in the \textit{supplementary material and video}. The video demonstrates STEP across diverse datasets, including animals~\cite{yang2022apt}, humans~\cite{doering2022posetrack21}, and laboratory species~\cite{lauer2022multi}, highlighting its robustness to variations in appearance, motion, and scene complexity. It further showcases STEP* with newly entering objects, Gaussian Maps predicted by GMSP, representative failure cases, and a real-world case study using live feeds from the Awajishima Monkey Center in Japan (recordings used with permission; live stream: \href{https://www.youtube.com/live/lsxYH2XQQCg}{\textcolor{blue}{Link}}). It also presents extensive quantitative analyses of detector-based initialization, initialization robustness, identity preservation, memory update, and sensitivity of pose and tracking performance to the memory confidence threshold. We refer the reader to the \textit{supp. material} for more details.

\subsection{Limitations}
\label{sec:lim}
STEP requires a box initialization in the first frame to establish the target state. While this can be automated using detectors~\cite{lyu2022rtmdet,zhang2022dino}, the approach still depends on the quality of the initialization.
We observe a coupling between tracking and pose estimation: inaccuracies in one task can propagate to the other. In particular, errors in box prediction can affect keypoint localization, and vice versa. This is reflected in the performance gap between \textbf{Ours} and \textbf{Ours*} (Table~\ref{tab:sota}), where removing tracking errors improves pose accuracy.

\section{Conclusion}
We propose STEP, a unified framework for simultaneous tracking and pose estimation across diverse species and domains. By leveraging a Transformer-based target-conditioned model-weight prediction paradigm, STEP jointly performs localization and keypoint estimation with only bounding-box initialization.
A key contribution of our approach is the integration of the GMSP and OMRA modules, which eliminates the need for explicit keypoint initialization while improving pose estimation performance. Unlike traditional top-down pipelines that rely on per-frame detections, STEP achieves efficient inference by maintaining temporal consistency through tracking.
Extensive experiments demonstrate state-of-the-art performance in pose estimation, competitive tracking accuracy, and robustness across challenging real-world scenarios. These results highlight STEP as a practical and scalable solution for joint tracking and pose estimation.

\section*{Acknowledgment}
This work was supported by the Dr. Vilas Mujumdar Chair at IIT Gandhinagar and, in part, by the Intramural Research Program of the National Institutes of Health (NIH). The contributions of the NIH authors are considered Works of the United States Government. The findings and conclusions presented in this paper are those of the authors and do not necessarily reflect the views of the NIH or the U.S. Department of Health and Human Services.
{
    \small
    \bibliographystyle{ieeenat_fullname}
    \bibliography{main}
}
\clearpage
\twocolumn[
\begin{center}
    {\LARGE \textbf{Supplementary Material}}
\end{center}
\vspace{1em}
]
\section{First Frame Initialization}

\subsection{Automatic Initialization.}
STEP requires only a bounding box in the first frame to initialize the target. To evaluate the robustness of this requirement, we replace manual initialization with detections from a pre-trained detector~\cite{zhang2022dino} on the CrowdPose validation set. Since detector outputs may include false positives or unmatched instances, we associate predicted boxes with ground truth using maximum IoU.
As shown in Table~\ref{tab:detman}, detector-based initialization results in minimal performance degradation compared to manual initialization. This demonstrates that STEP can be effectively deployed in a fully automated setting without requiring manual intervention.
Importantly, while initialization remains necessary, it is a weak constraint in practice. In controlled environments (e.g., laboratory or zoo settings), object localization is relatively straightforward, and simple detection can be readily integrated as initialization. This makes STEP applicable across both constrained and in-the-wild scenarios.

\begin{figure*}[ht]
    \begin{center}
    \includegraphics[width=\textwidth]{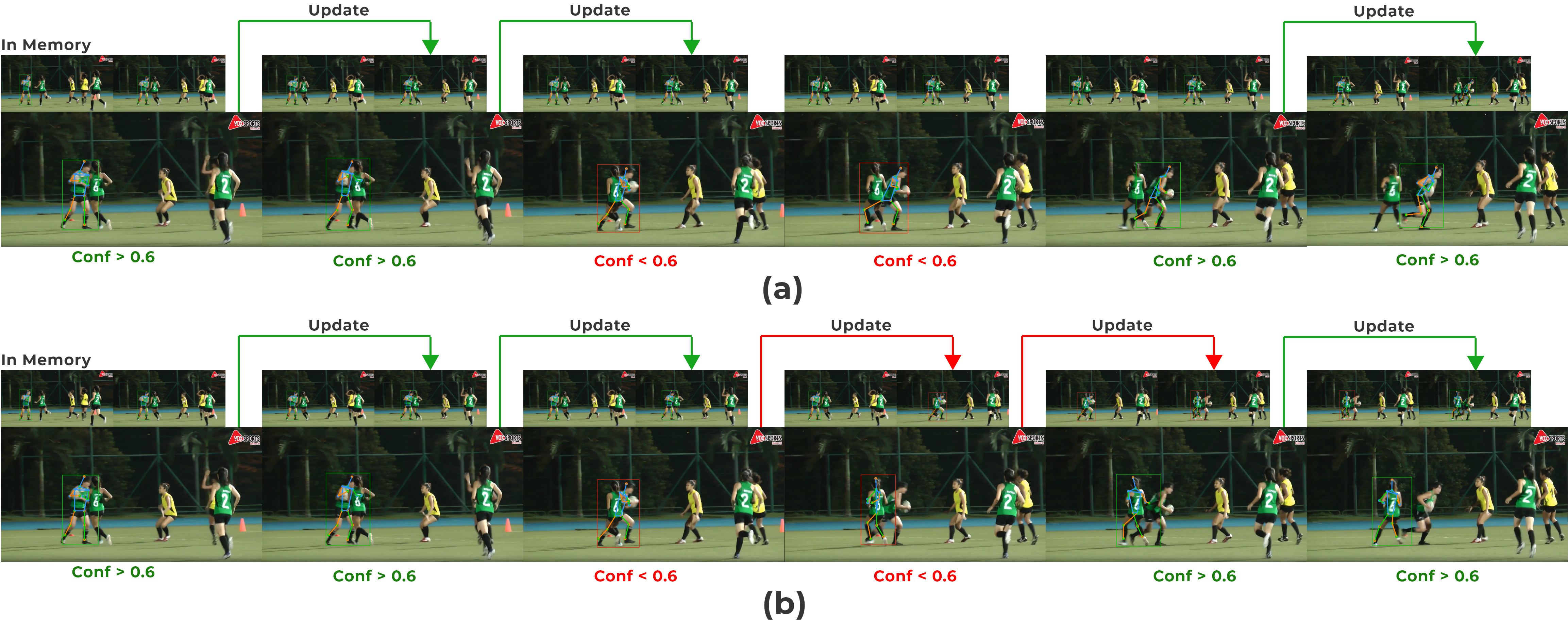}
    \end{center}
    \vspace{-0.5cm}
    \caption{We demonstrate tracking and pose estimation during the occlusion of a target human instance by another human with a similar appearance. Zoom in is recommended for clarity. Panel (a) illustrates memory updates occurring only when at least 50\% of keypoints and the bounding box localization are predicted with a probability greater than $\tau_m = 0.6$. In contrast, panel (b) shows memory updates being performed regardless of the probability value. In both panels, low-confidence frames, which may introduce or propagate errors, are indicated by a red bounding box. In panel (b), STEP wrongly adjusts its confidence from the original target to the new instance based on the memory state. Subsequently, it starts tracking and performing pose estimation on the new object. 
}
\label{fig:mu}
\vspace{-0.3cm}
\end{figure*}

\begin{figure*}[t]
    \centering
    \begin{minipage}[t]{0.34\linewidth}
        \centering
        \includegraphics[width=\linewidth]{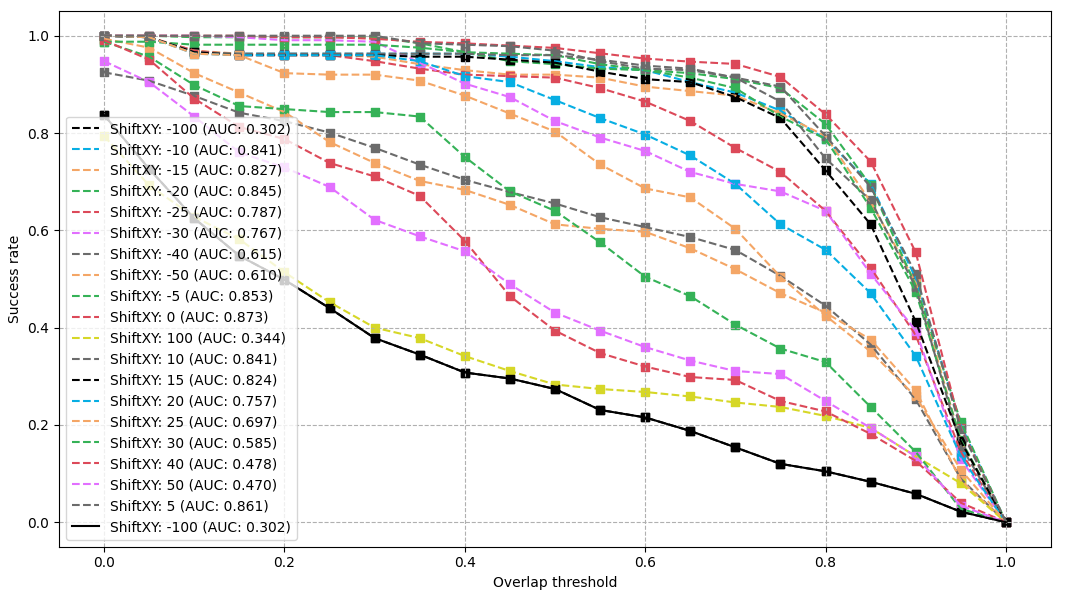}
        \vspace{-5pt}
    \end{minipage}
    \hfill
    \begin{minipage}[t]{0.32\linewidth}
        \centering
        \includegraphics[width=\linewidth]{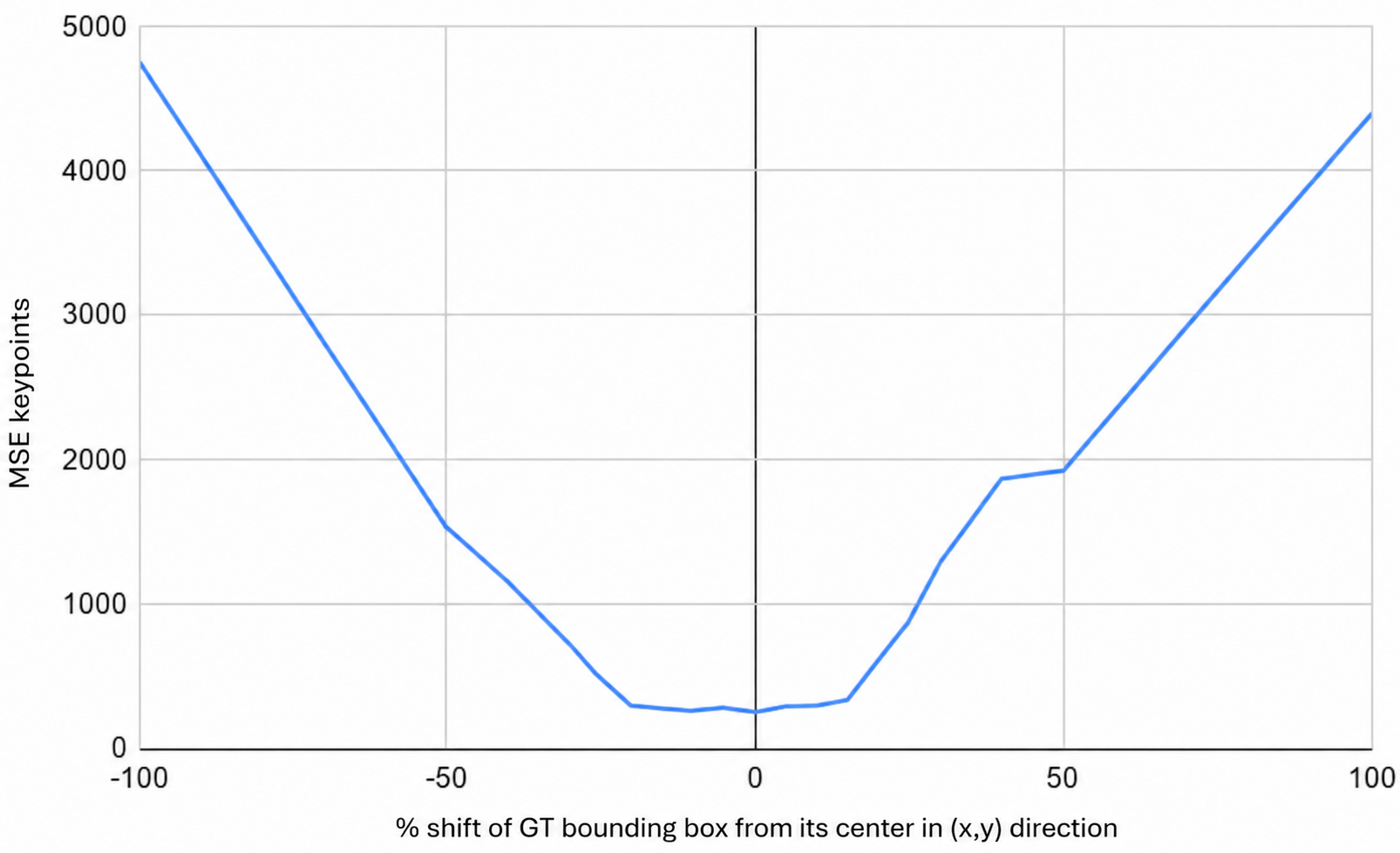}
        \vspace{-5pt}
    \end{minipage}
    \hfill
    \begin{minipage}[t]{0.32\linewidth}
        \centering
        \includegraphics[width=\linewidth]{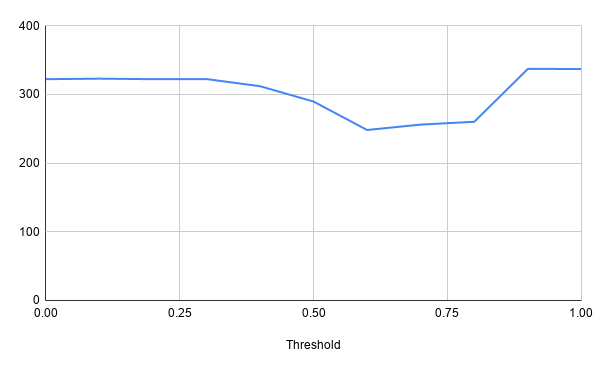}
        \vspace{-5pt}
    \end{minipage}
    \vspace{-2pt}
    \caption{
    \textbf{Left:} Robustness to noisy first-frame initialization.
    \textbf{Middle:} Influence of tracking quality on pose estimation.
    \textbf{Right:} Sensitivity to the memory confidence threshold $\tau$.
    (Please zoom for details.)
    }
    \label{fig:robustness_threshold}
    \vspace{-0.2cm}
\end{figure*}

\subsection{Robustness to Initialization Errors.} Since STEP jointly performs tracking and pose estimation, tracking errors can affect the subsequent pose predictions. To quantify this effect, we intentionally perturb the first-frame bounding-box initialization by varying amounts in the $x$ and $y$ directions, expressed as percentages of the box width and height, respectively. Figure ~\ref{fig:robustness_threshold} (Left and Middle) shows the resulting tracking and pose performance. STEP remains largely robust to moderate localization errors; however, both tracking and pose accuracy degrade noticeably under large initialization shifts. These results indicate that substantial tracking failures are a primary source of degraded pose estimation performance, while moderate initialization inaccuracies have a limited impact.

\subsection{Memory Update}
Memory updates lead to consistent performance improvements by adapting the target representation over time, thereby improving robustness to appearance changes and mitigating drift. Figure~\ref{fig:mu} presents a qualitative example illustrating the effectiveness of memory updates. By continuously adapting target representations, DeepSTEP maintains stable tracking and accurate pose estimation despite appearance variations, occlusions, and other challenging conditions.

\begin{table}[t]
\scriptsize
\setlength{\tabcolsep}{4pt}
\renewcommand{\arraystretch}{1.1}
\begin{center}
\begin{tabular}{l c c c c c}
\toprule
\textbf{Model} & \textbf{Initialization} & \textbf{Per Frame Box} & \textbf{MSE $\downarrow$} & \textbf{OKS $\uparrow$} & \textbf{PDJ@0.08 $\uparrow$} \\
\midrule

STEP  & Ground Truth & $\times$ & 292.86 & 0.734 & 0.861 \\
STEP* & Ground Truth & $\checkmark$ & 214.87 & 0.736 & 0.891 \\
STEP  & Detector     & $\times$ & 292.31 & 0.731 & 0.835 \\
STEP* & Detector     & $\checkmark$ & 213.62 & 0.733 & 0.854 \\

\bottomrule
\end{tabular}
\end{center}
\caption{Initialization using a detector \cite{zhang2022dino} vs GT boxes.}
\label{tab:detman}
\vspace{-0.4cm}
\end{table}

\begin{table}[t]
\scriptsize
\renewcommand{\arraystretch}{1.05}
\begin{center}
\begin{tabular}{c|cccccc}
\hline
$\tau$ & \textbf{DiMP-18} & \textbf{ATOM} & \textbf{ToMP-50} & \textbf{GMOT} & \textbf{TaMOs} & \textbf{STEP} \\
\hline
0.05 & 0.997 & 0.993 & 0.999 & 0.998 & 0.998 & 0.998 \\
0.15 & 0.997 & 0.990 & 0.999 & 0.999 & 0.999 & 0.998 \\
0.30 & 0.988 & 0.981 & 0.998 & 0.998 & 0.992 & 0.993 \\
0.45 & 0.971 & 0.954 & 0.991 & 0.989 & 0.985 & 0.987 \\
0.60 & 0.928 & 0.900 & 0.967 & 0.964 & 0.967 & 0.962 \\
0.75 & 0.779 & 0.753 & 0.866 & 0.878 & \textbf{0.914} & 0.891 \\
0.90 & 0.300 & 0.257 & 0.558 & 0.611 & \textbf{0.682} & 0.664 \\
0.95 & 0.094 & 0.090 & 0.249 & 0.331 & 0.384 & \textbf{0.420} \\
\hline
\end{tabular}
\end{center}
\caption{\textbf{Tracking.} Overlap Precision (OP$_{\tau}$) at different IoU thresholds for STEP compared with DiMP \cite{bhat2019learning}, ATOM \cite{danelljan2019atom}, ToMP \cite{mayer2022transforming}, GMOT \cite{verma2025gmot}, and TaMOs \cite{mayer2024beyond}.}
\vspace{-0.4cm}
\label{tab:conf_threshold}
\end{table}

\section{Quantitative Analysis}
\begin{figure*}[t]
    \centering
    \includegraphics[width=0.9\linewidth ]{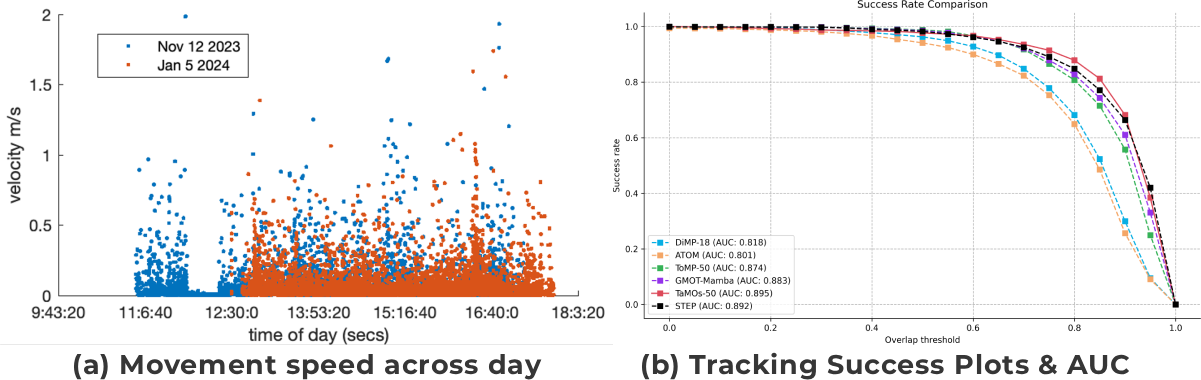}
    \caption{(a) Movement velocity observed on recorded data from AMC (tracks $\geq 1$ sec). Distances are derived from AMC's perspective on Google Maps. (b) IoU and Overlap Precision $(\text{OP}_T)$ for bounding box tracking in STEP compared with: DiMP \protect\cite{bhat2019learning}, ToMP \protect\cite{mayer2022transforming}, ATOM \protect\cite{danelljan2019atom}, TAMOS \protect\cite{mayer2024beyond}, and GMOT-Mamba \protect\cite{verma2025gmot}.
    }
    \label{fig:tracking_supp}
    \vspace{-0.2cm}
\end{figure*}

\noindent \textbf{1. Tracking and Identity Preservation.}
Table~\ref{tab:conf_threshold} and Figure~\ref{fig:tracking_supp}(b) compare our tracking performance with state-of-the-art methods across different values of $\tau$.
To further assess STEP's ability to preserve object identities, we evaluate it on a subset of the validation set containing videos with more than two objects. This subset comprises 81 videos, 1,022 frames, and 184 tracks. Using an IoU threshold of $0.6$, STEP achieves an IDF1 score of $0.858$, with 128 ID switches, 69 track fragmentations, and 27 unrecovered track losses. These results demonstrate that STEP maintains strong identity consistency despite operating without per-frame detections.

\begin{figure}[t]
    \centering
    \includegraphics[width=0.95\columnwidth]{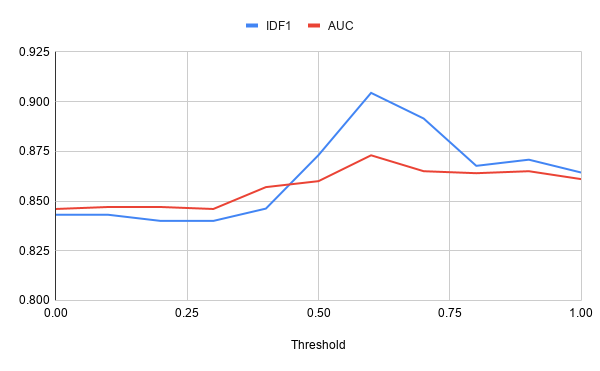}
    \vspace{-6pt}
    \caption{\textbf{Sensitivity to the memory confidence threshold $\tau$.} Effect of $\tau$ used for memory updates on tracking performance in terms of IDF1 and AUC.}
    \label{fig:conf_thresh_track}
    \vspace{-8pt}
\end{figure}

\noindent \textbf{2. Sensitivity to the Confidence Threshold.}
We analyze the sensitivity of STEP to the confidence threshold $\tau$ used for memory updates. Specifically, we vary $\tau$ over the range $[0,1]$ and evaluate the resulting pose and tracking performance, as shown in Figure~\ref{fig:robustness_threshold} (Right) and Figure \ref{fig:conf_thresh_track}. The results demonstrate the effect of the threshold on the reliability of memory updates and overall tracking and pose estimation performance.

\noindent \textbf{3. Comparison with Species-Specific Baselines.}
Table \ref{tab:macaque_horse} presents additional evaluation on macaque and horse subsets from APT36K, containing approximately 930 frames with 77 instances and 285 frames with 28 instances, respectively. DeepLabCut models trained on external macaque \cite{labuguen2021macaquepose} and horse \cite{mathis2021pretraining} datasets exhibit limited generalization on these subsets, likely due in part to domain differences between the training and evaluation data. Nevertheless, the performance gap remains substantial across both subsets.

\begin{figure*}[t]
    \centering
    \includegraphics[width=\linewidth ]{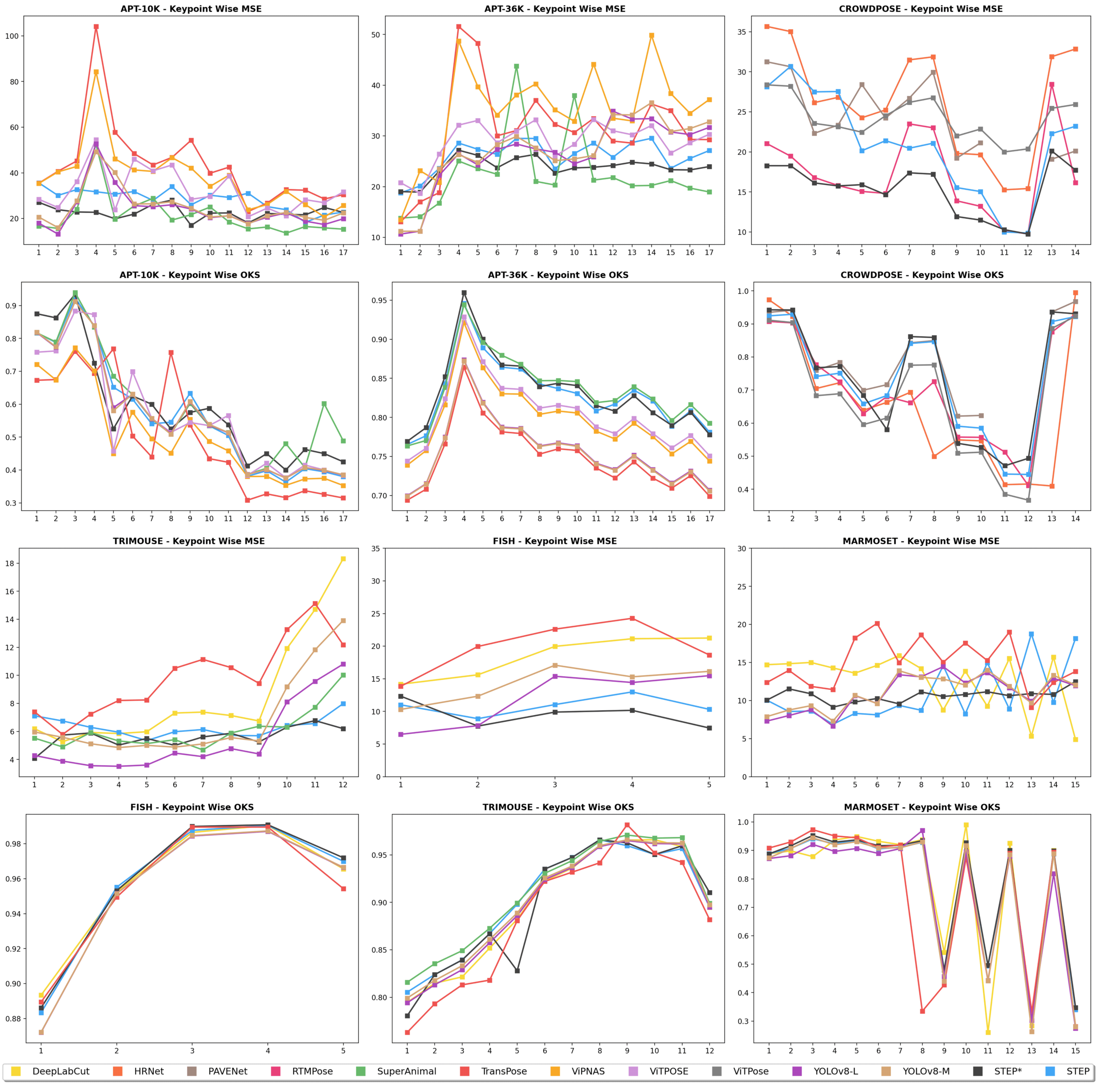}
    \vspace{-0.1cm}
    \caption{Comparison of Mean Squared Error (MSE) and Object Keypoint Similarity (OKS) for individual keypoints against various existing methods: VitPose \protect\cite{xu2022vitpose}, YOLOv8 \protect\cite{jocher2023yolo}, TransPose \protect\cite{yang2021transpose}, ViPNAS \protect\cite{xu2021vipnas}, DeepLabCut \protect\cite{lauer2022multi}, SuperAnimal \cite{ye2024superanimal}, HRNet \protect\cite{sun2019deep}, RTMPose \protect\cite{rtmpose}, and PAVENet \cite{yu2026end}. Notably, the marmoset dataset features top-view images of marmosets in a cage with infrequent occurrences of lower limb keypoints. The inherent imbalance in training data leads to notable inaccuracies across all methods for the four bottom-limb keypoints}
    \label{fig:kpserr}
    \vspace{-0.2cm}
\end{figure*}

\noindent \textbf{4. Per keypoint Metrics.}
Corresponding to the consolidated keypoint metrics reported in Table 1 of the main paper, Figure \ref{fig:kpserr}presents the performance metrics for each individual keypoint.

\begin{figure*}[t]
    \centering
    \includegraphics[width=\linewidth ]{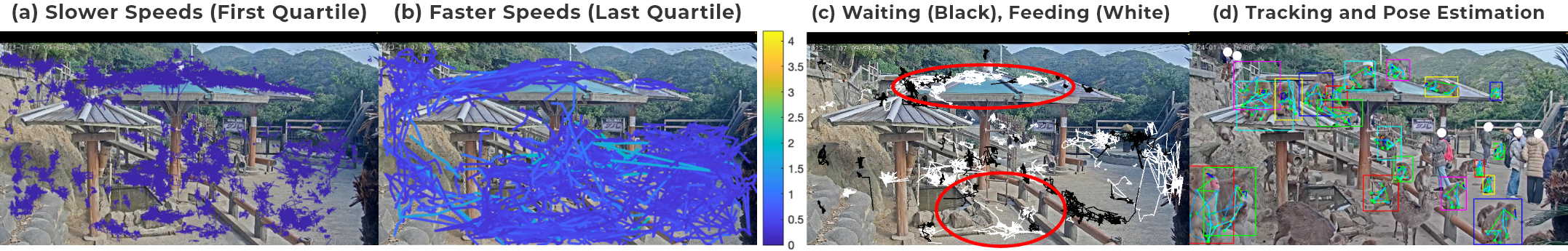}
    \vspace{-0.5cm}
    \caption{(a) Slowest and (b) fastest quartiles of self-movement velocities over two days of recordings. Feeding events cause movement toward food locations, resulting in transitions from black to white clusters, shown in (c). Trajectories are obtained from bounding box centers. A high-occupancy scene with overlaid tracks and poses is shown in (d).
    }
    \label{fig:awazip}
\end{figure*}

\section{A Case Study on AMC Live Streams.}
To evaluate STEP in a real-world setting, we apply it to live video streams from the Awajishima Monkey Center in Japan (\noindent Video recordings used with permission; Live Stream: \href{https://www.youtube.com/live/lsxYH2XQQCg}{\textcolor{blue}{Link}}). The center houses a large group of freely moving macaques, which are captured on a live camera feed overlooking a popular visitor area.
Using detector-based initialization~\cite{lyu2022rtmdet}, STEP successfully tracks individuals and estimates pose across frames, enabling the extraction of fine-grained behavioral signals. For example, we observe that monkeys typically exhibit low median speeds (below $1$ m/s), with bursts of high activity during feeding events (Figure~\ref{fig:awazip}). Temporal analysis (Figure~\ref{fig:tracking_supp}(a)) further reveals variations in movement patterns across individuals and time.
Although analysis is limited to 2D image space, these results demonstrate that STEP can serve as a practical tool for large-scale behavioral analysis in unconstrained environments, providing insights beyond coarse activity measurements.

\section{Qualitative Results.}
\begin{figure*}[t]
    \centering
    \includegraphics[width=\linewidth ]{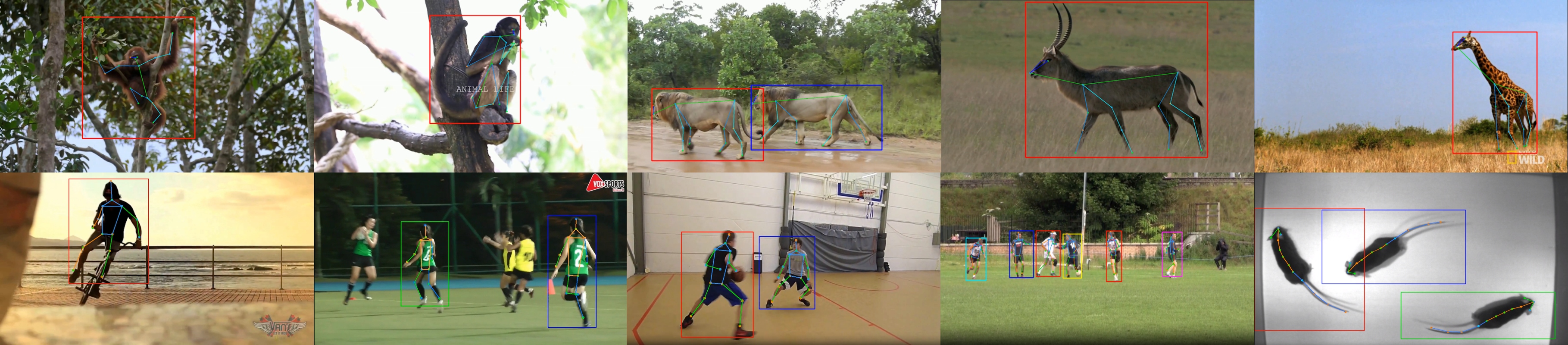}
    \vspace{-0.5cm}
    \caption{Frames from supplementary video showcasing tracking and pose estimation. Please refer to the supplementary video for qualitative results on video sequences.
    }
    \vspace{-0.2cm}
    \label{fig:ah}
\end{figure*}

Frames from the supplementary video that showcase tracking and pose estimation results are shown in Figure~\ref{fig:ah}. Please refer to the accompanying supplementary video for qualitative results across diverse video sequences, demonstrating the effectiveness of our method in tracking and pose estimation.

\begin{figure*}[t]
    \centering
    \includegraphics[width=\linewidth ]{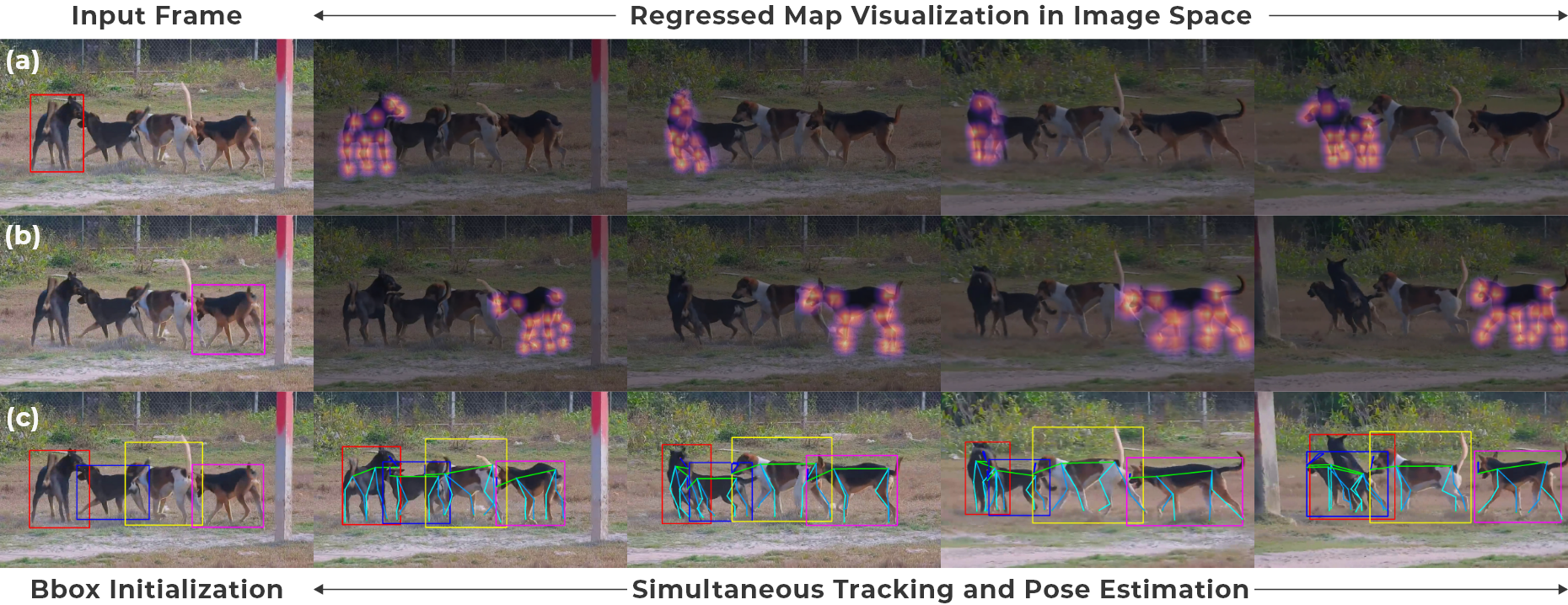}
    \vspace{-0.2cm}
    \caption{
    For a video sequence, columns display target score maps $\hat{K}_\text{gm}$ when initiated with bounding boxes, as depicted in (a) and (b).  In (c), STEP runs concurrently for each bounding box to track and estimate the pose of respective targets. Notably, observe STEP's robustness in estimating keypoints, particularly in occluded scenarios.
    A failure case is showcased in the last frame of (c) for the red bounded target, stemming from ambiguous activations within the red target's score map, as evident in the last frame of (a).
    }
    \label{fig:mapvis}
    \vspace{-0.2cm}
\end{figure*}

\noindent \textbf{GMSP Map Visualization}
Figure~\ref{fig:mapvis} visualizes the target score maps generated by STEP during tracking and pose estimation. These visualizations provide insights into the model's localization and keypoint estimation behavior, particularly under occlusion and in challenging cases. The figure also illustrates a failure case caused by ambiguous score-map activations.

\begin{table}[t]
\scriptsize
\caption{
Evaluation on the primate and horse subsets.
}
\vspace{-0.4cm}
\label{tab:macaque_horse}
\renewcommand{\arraystretch}{1.05}
\begin{center}
\begin{tabular}{lcc lcc}
\toprule
\multicolumn{3}{c}{\textbf{Primates (APT36K)}} &
\multicolumn{3}{c}{\textbf{Horses (APT36K)}} \\
\cmidrule(lr){1-3}
\cmidrule(lr){4-6}
Model & MSE $\downarrow$ & OKS $\uparrow$
& Model & MSE $\downarrow$ & OKS $\uparrow$ \\
\midrule
\rowcolor{ourslight}
STEP & \cellcolor{best}{437.15} & \cellcolor{best}{0.783}
& STEP & \cellcolor{best}{503.63} & \cellcolor{best}{0.784} \\
DLC-Macaque & 1948.82 & 0.650
& DLC-Horse & 3833.83 & 0.764 \\

\bottomrule
\end{tabular}
\end{center}
\vspace{-0.4cm}
\end{table}

\end{document}